# One-Step Detection Paradigm for Hyperspectral Anomaly Detection via Spectral Deviation Relationship Learning

Jingtao Li, Xinyu Wang, Shaoyu Wang, Hengwei Zhao, Liangpei Zhang, *Fellow, IEEE,* and Yanfei Zhong, *Senior Member, IEEE*

*Abstract*—**Hyperspectral anomaly detection (HAD) involves identifying the targets that deviate spectrally from their surroundings, without prior knowledge. Recently, deep learning based methods have become the mainstream HAD methods, due to their powerful spatial-spectral feature extraction ability. However, the current deep detection models are optimized to complete a proxy task (two-step paradigm), such as background reconstruction or generation, rather than achieving anomaly detection directly. This leads to suboptimal results and poor transferability, which means that the deep model is trained and tested on the same image. In this paper, an unsupervised transferred direct detection (TDD) model is proposed, which is optimized directly for the anomaly detection task (one-step paradigm) and has transferability. Specially, the TDD model is optimized to identify the spectral deviation relationship according to the anomaly definition. Compared to learning the specific background distribution as most models do, the spectral deviation relationship is universal for different images and guarantees the model transferability. To train the TDD model in an unsupervised manner, an anomaly sample simulation strategy is proposed to generate numerous pairs of anomaly samples. Furthermore, a global self-attention module and a local self-attention module are designed to help the model focus on the "spectrally deviating" relationship. The TDD model was validated on four public HAD datasets. The results show that the proposed TDD model can successfully overcome the limitation of traditional model training and testing on a single image, and the model has a powerful detection ability and excellent transferability.**

## I. Introduction

Due to the rich spectral information, hyperspectral imagery (HSI) can be used to detect anomalies with little spatial information [1]. Hyperspectral anomaly detection (HAD) is aimed at detecting pixels that deviate spectrally from the surroundings [2], and has been proven valuable in many areas, including infected tree detection [3], rare mineral detection [4], and defense application. Generally speaking, anomalies occupy a low proportion of the image and refer to man-made targets, natural objects, and other interferers [3].

The current HAD models can be divided into three main categories: 1) statistics-based models; 2) representation-based models; and 3) deep learning based models. The statistics-based models for anomaly detection involve solving a binary hypothesis testing problem [3]. Because the anomaly distribution is unknown, these models construct the test statistics by assuming that the background conforms to a certain statistical distribution [5]–[7]. For example, the famous Reed-Xiaoli (RX) detector assumes that the background obeys a Gaussian distribution, and the Mahalanobis distance between the test pixel and the obtained distribution is used to measure the anomaly degree [8]. However, despite the high detection efficiency, the assumed distribution may not hold in complex backgrounds and can lead to unsatisfactory detection results [9]. The representation-based models mostly detect anomalies by exploiting the low-rank characteristic of the background and the sparse characteristic of the anomalies [10]–[12]. The obtained sparse component then represents the anomalies. Some other representation-based models employ tensor representation to reflect the three-dimensional (3D) structure of the HSI, and can achieve superior detection results [13]. However, a suitable background dictionary plays a key role in these methods, but this can be difficult to construct without any prior knowledge in real applications [14].

In recent years, deep learning based methods have become the mainstream methods because of their powerful spatial-spectral feature extraction ability [15], [16]. However, the current deep learning based HAD models follow the two-step detection paradigm In the first step, the model learns the background distribution via a proxy task, such as background reconstruction [17]–[20]. In the second step, the anomalies are identified by measuring the difference between the input image and the reconstructed background. This two-step detection

This work was supported by the National Key Research and Development Program of China under Grant No. 2022YFB3903500, in part by the National Natural Science Foundation of China under Grant No. 42101327 and No. 42071350, and in part by LIESMARS Special Research Funding. *(Corresponding author: Xinyu Wang.).*

Jingtao Li, Shaoyu Wang, Hengwei Zhao, Liangpei Zhang, and Yanfei Zhong are with the State Key Laboratory of Information Engineering in Surveying, Mapping and Remote Sensing and the Hubei Provincial Engineering Research Center of Natural Resources Remote Sensing Monitoring, Wuhan University, Wuhan 430072, China (e-mail: jingtaoli@whu.edu.cn; wangshaoyu@whu.edu.cn; whu_zhaohw@whu.edu.cn; zlp62@whu.edu.cn; zhongyanfei@whu.edu.cn).

Xinyu Wang is with the School of Remote Sensing and Information Engineering, Wuhan University, Wuhan 430072, China (e-mail: wangxinyu@whu.edu.cn).

paradigm leads to two serious problems. 1) A proxy task may obtain suboptimal results because the basic assumption of the proxy task for HAD may not hold [21]. The auto-encoder (AE) assumes that the background is more easily reconstructed than the anomalies. Unfortunately, the reconstruction ability of the AE may be migrated to the anomalies, especially when the background distribution is complex [22]. Similarly, generative adversarial networks (GANs) consider that the background can be generated more easily than anomalies, but GAN models may also generate samples that are out of the normal background manifold [21]. 2) The trained model lacks transferability to different images because it aims to learn a certain background distribution in the training stage [23]. The background distribution varies in different images, which hinders the transferability. Thus the deep learning based HAD models are trained and tested on the same image [1], [15]. Although Li *et al*. [24] attempted to give a HAD model transferability, the HAD model was trained in a supervised manner, and was not suitable for a real application.

In this paper, a *transferred direct* detection (TDD) model is proposed for the HAD task. Fig. 1 shows the difference between the TDD model and the traditional deep model. The TDD model outputs the anomaly map *directly*, rather than the reconstructed background, which is called the "one-step detection paradigm" in this paper. The TDD model is optimized to identity a "spectrally deviating" relationship according to the anomaly definition, rather than a specific background distribution, generating the *transferability* for different images. In this study, the TDD model was validated on four public HAD datasets, where the Hyperspectral Digital Imagery Collection Experiment (HYDICE) dataset was used for unsupervised training and the remaining datasets were inferred directly, without further fine-tuning. The results show that the proposed TDD model can successfully overcome the limitation of traditional model training and testing on a single image, and has a powerful detection ability and excellent transferability.

The main contributions of this paper can be summarized as follows.

1) The one-step detection paradigm for the HAD task is proposed and implemented (i.e., the TDD model). Differing from the traditional two-step paradigm, the TDD model is optimized *directly* for the HAD task and has good *transferability*. To the best of our knowledge, the TDD model is the first unsupervised HAD model with transferability.
2) An anomaly sample simulation strategy is proposed to train the TDD model in an unsupervised manner. The generated anomaly samples then optimize the TDD model *directly* for the HAD task.
3) A global self-attention module (GAM) and a local self-attention module (LAM) are designed to help the model focus on a "spectrally deviating" relationship, rather than a specific background, enhancing the detection transferability.

The rest of this paper is organized as follows. Section II introduces the current algorithms for the HAD task. Section III describes the one-step detection paradigm and the TDD model.

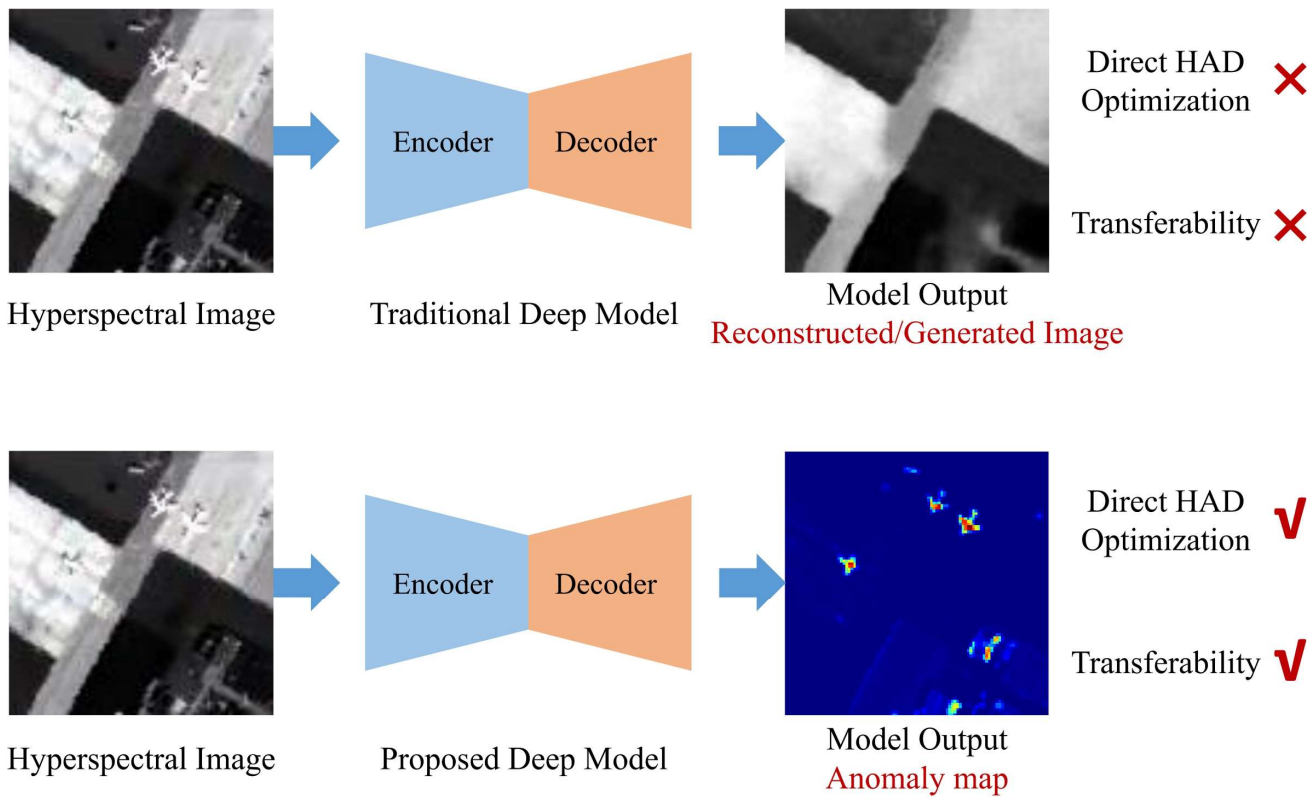

**Fig. 1.** Comparison between the two-step paradigm of the traditional model and the proposed one-step paradigm for model training.

Section IV presents the comparative results and the model analysis. Finally, Section V concludes the paper.

## II. Related Work

### A. Statistics-Based Models

The statistics-based detection models assume that the background conforms to a certain statistical distribution [25], and the distribution density implies the anomaly degree [3]. The RX detector [8] is a milestone statistical method which assumes that the background obeys a Gaussian distribution, and the Mahalanobis distance between the test pixel and the obtained distribution is used to measure the anomaly degree. Inspired by the classical RX detector, a series of extensions have been proposed. For example, Schaum [6] used the principal components of the background covariance to conduct HAD in the subspace, to obtain a robust performance. Similarly, the kernel RX algorithm [26] was proposed to detect anomalies in the high-dimensional feature space. Guo *et al*. [27] proposed weighted-RX and linear filter based RX detectors, to achieve better background estimation. Some other traditional detectors are also based on statistical modeling, such as the manifold learning detector [17] and the support vector based detector [28]. However, the statistics-based methods always make various distribution assumptions, such as the assumption of a Gaussian distribution, which may not hold in complex backgrounds and can lead to unsatisfactory detection results [29].

### B. Representation-Based Models

The representation-based models detect anomalies using some of the HSI properties, such as the low-rank characteristic of the background or the sparse characteristic of the anomalies [14]. The obtained sparse component then represents the anomalies. Low-rank and sparse matrix decomposition (LSDM) [12] has been successfully applied for the HAD task, by decomposing the HSI into a low-rank background and sparse anomalies. Chen *et al*. [11] implemented the LSDM technique with the robust principal component analysis (RPCA) algorithm. Zhang *et al*. [30] focused more on the low-rank prior and proposed a new Mahalanobis distance based detector. Cheng and Wang [31] incorporated the spatial information into a low-rank model with graph regularization and total variation regularization. More recently, some researchers have employed tensor representation to reflect the 3D structure of the HIS [32]. For example, Li *et al*. [13] developed a prior-based tensor approximation (PTA) method, which combines low-rank, sparse, and piecewise smooth priors with the advantages of tensor representation. However, the above representation-based methods rely on hand-crafted priors and a constructed background dictionary, and have a limited ability to characterize the real background in HSI [14].

### C. Deep Learning Based Models

The deep learning based methods always assume that the background can be reconstructed better than the anomalies. They follow the two-step detection paradigm, where the first step involves training a deep reconstruction model and the second step involves outputting the detection map using the reconstructed background [15]. Li *et al*. [24] were the first to introduce a convolutional neural network (CNN) into HAD and detected anomalies in a supervised manner. Wang *et al*. [15] proposed an autonomous hyperspectral anomaly detection network (Auto-AD), in which the background is reconstructed by the AE and the anomalies appear as reconstruction errors. An adaptive-weighted loss function was also designed to further suppress the anomaly reconstruction. Xie *et al*. [18] proposed a spectral constrained adversarial AE (SC_AAE) to perform background suppression and discriminative representation extraction. Wang *et al*. [1] designed a deep low-rank prior based method (DeepLR), which combines a model-driven low-rank prior and a data-driven AE. Li *et al*. [19] developed a sparse coding (SC)-inspired GAN for weakly supervised HAD, which learns a discriminative latent reconstruction with small errors for background pixels and large errors for anomalous ones. Arisoy *et al*. [20] trained a GAN model to generate a synthetic background image which is close to the original background image. Despite the excellent performance, the two-step detection paradigm can cause suboptimal results and poor transferability (as discussed in the Introduction section).

## III. PROPOSED MODEL

The TDD model is proposed in this paper to implement the one-step detection paradigm, where the model outputs the anomaly map directly and can be transferred to different hyperspectral images (as shown in Fig. 2). The spectral deviation relationship is focused in one-step paradigm rather than the certain background in traditional two-step paradigm. In this section, we first formulate the one-step detection paradigm, and then introduce the designed anomaly sample simulation strategy, i.e., the GAM and LAM (as described in the Model Architecture section). The training loss and the model transferability are then discussed.

### A. One-Step Detection Paradigm

Given a hyperspectral image $\boldsymbol{X} \in R^{H\times W\times B}$, where $H$, $W$, and $B$ are the height, width, and band number, respectively, the HAD task is a function $\phi$ mapping $\boldsymbol{X}$ to the estimated detection map $\hat{\boldsymbol{M}} \in R^{H\times W}$. $\boldsymbol{M}$ is the ground truth. $\boldsymbol{X} = \boldsymbol{BKG} + \boldsymbol{A}$, where $\boldsymbol{BKG}$ is the background component and $\boldsymbol{A}$ is the anomaly component. The value of $\hat{\boldsymbol{M}}$ represents the anomaly degree of the corresponding pixel.

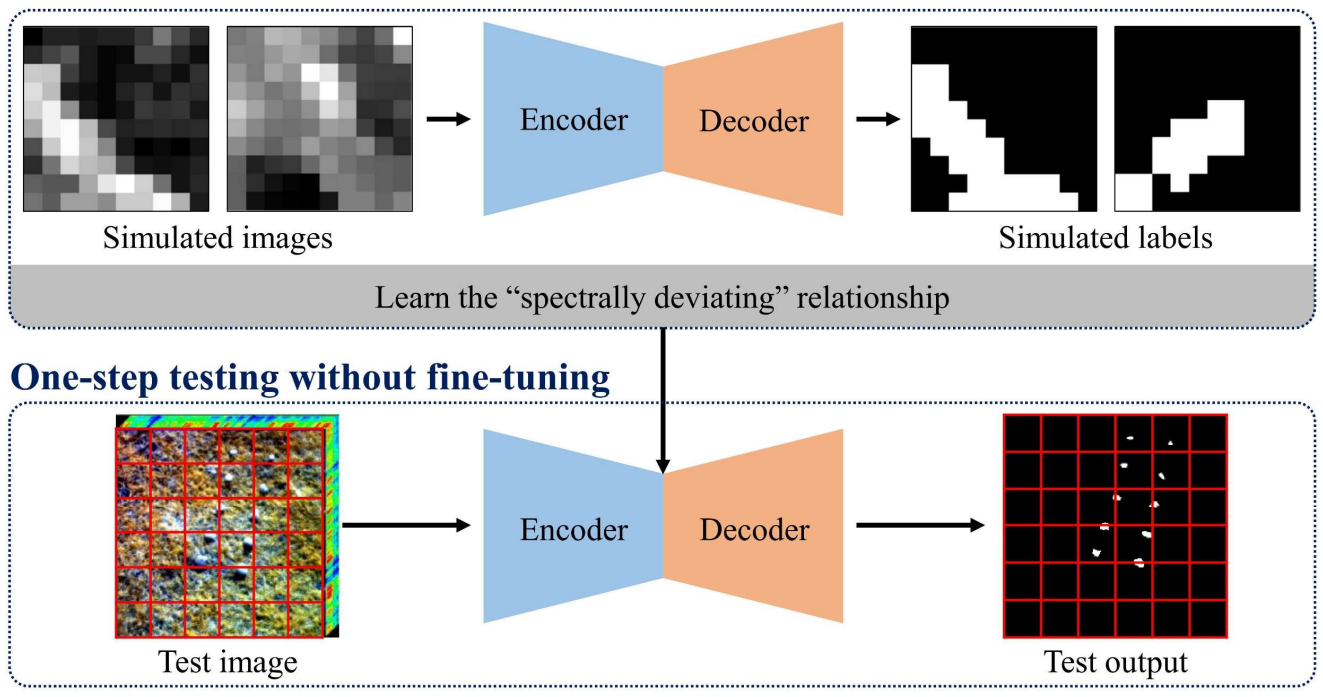

**Fig. 2.** Overview of the TDD model, which implements the proposed one-step detection paradigm. The TDD model is trained using simulated sample pairs and has the ability to infer the test image without any fine-tuning.

The traditional deep HAD models separate $\phi$ into two steps: $\phi_1$ and $\phi_2$. The mappings of $\phi_1$ and $\phi_2$ are shown in (1)–(2). $\phi_1$ always adopts a proxy task and outputs a reconstruction version $\boldsymbol{X}'$. $\boldsymbol{X}'$ is expected to only reconstruct the background. $\phi_2$ then computes the difference between $\boldsymbol{X}$ and $\boldsymbol{X}'$ using a certain metric to obtain $\hat{\boldsymbol{M}}$. $\phi_1$ learns the distribution of $\boldsymbol{BKG}$ and $\phi_2$ learns the distribution of $\boldsymbol{A}$. The model optimization objective is to minimize the difference between $\boldsymbol{X}$ and $\boldsymbol{X}'$ of $\phi_1$, as in (3). Although the two-step strategy can achieve satisfactory results, the adopted proxy task can lead to suboptimal results and poor transferability, as discussed previously.

$$\phi_1 : \boldsymbol{X} \to \boldsymbol{X}' \text{ (learn } \boldsymbol{BKG}\text{)} \tag{1}$$
$$\phi_2 : (\boldsymbol{X}, \boldsymbol{X}') \to \boldsymbol{M} \text{ (learn } \boldsymbol{A}\text{)} \tag{2}$$
$$Loss = \text{Compare } (\boldsymbol{X}, \boldsymbol{X}') \tag{3}$$

To overcome these limitations, we propose a one-step detection paradigm, where the model outputs $\hat{\boldsymbol{M}}$ from $\boldsymbol{X}$ directly, as in (4). The one-step model does not learn a specific distribution of $\boldsymbol{BKG}$ or $\boldsymbol{A}$, but learns the unified relationship $R(\boldsymbol{BKG}, \boldsymbol{A})$ between $\boldsymbol{BKG}$ and $\boldsymbol{A}$. For the HAD task, $R(\boldsymbol{BKG}, \boldsymbol{A})$ is the "spectrally deviating" relationship of $\boldsymbol{A}$ relative to $\boldsymbol{BKG}$. The model optimization objective directly minimizes the difference between $\hat{\boldsymbol{M}}$ and $\boldsymbol{M}$, as in (5). Compared to the two-step detection paradigm, the one-step detection paradigm has two main advantages, as shown in Fig. 1. 1) The model is optimized directly for the HAD task, without any post-processing process (i.e., $\phi_2$). 2) Transferability can be achieved since the model learns the unified $R(\boldsymbol{BKG}, \boldsymbol{A})$ rather than a specific distribution of $\boldsymbol{BKG}$ or $\boldsymbol{A}$. To the best of our knowledge, the TDD model is the first one-step detection model for the HAD task.

$$\phi : \boldsymbol{X} \to \hat{\boldsymbol{M}} \text{ (learn } R(\boldsymbol{BKG}, \boldsymbol{A})\text{)} \tag{4}$$
$$Loss = \text{Compare } (\hat{\boldsymbol{M}}, \boldsymbol{M}) \tag{5}$$

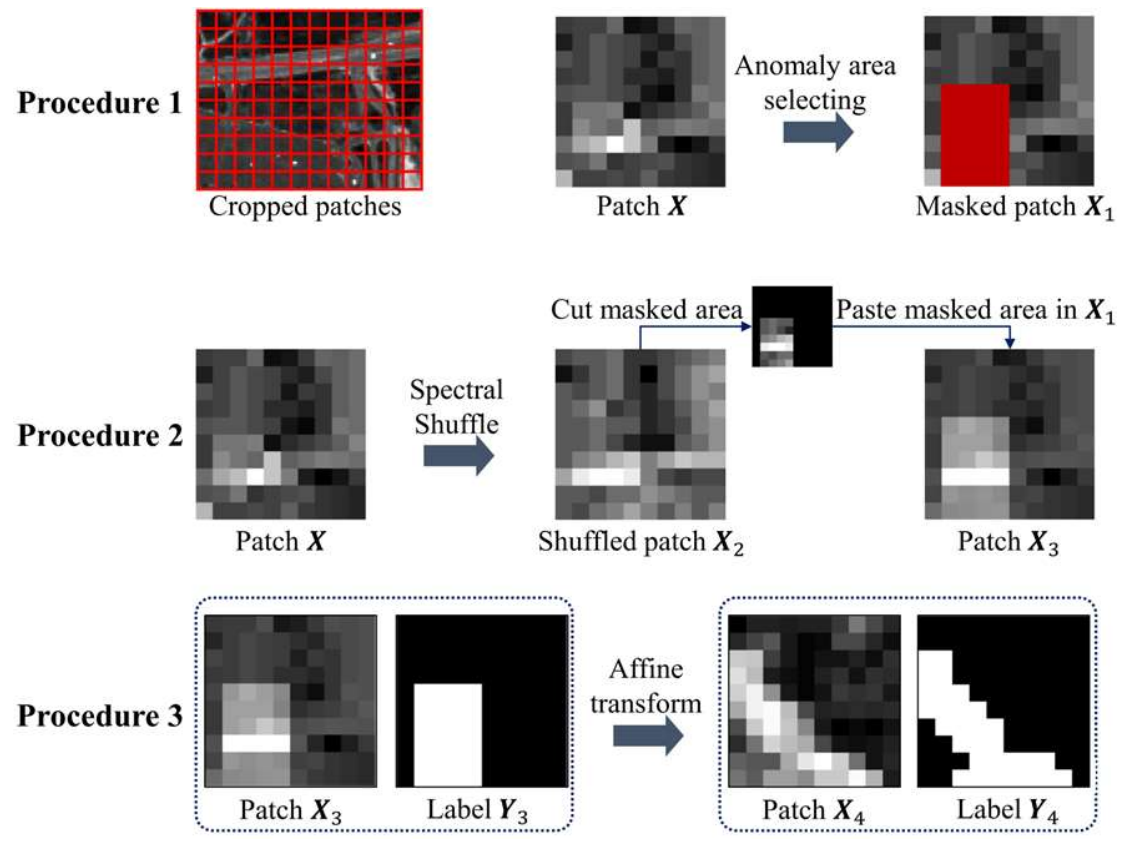

**Fig. 3.** Proposed anomaly sample simulation strategy.

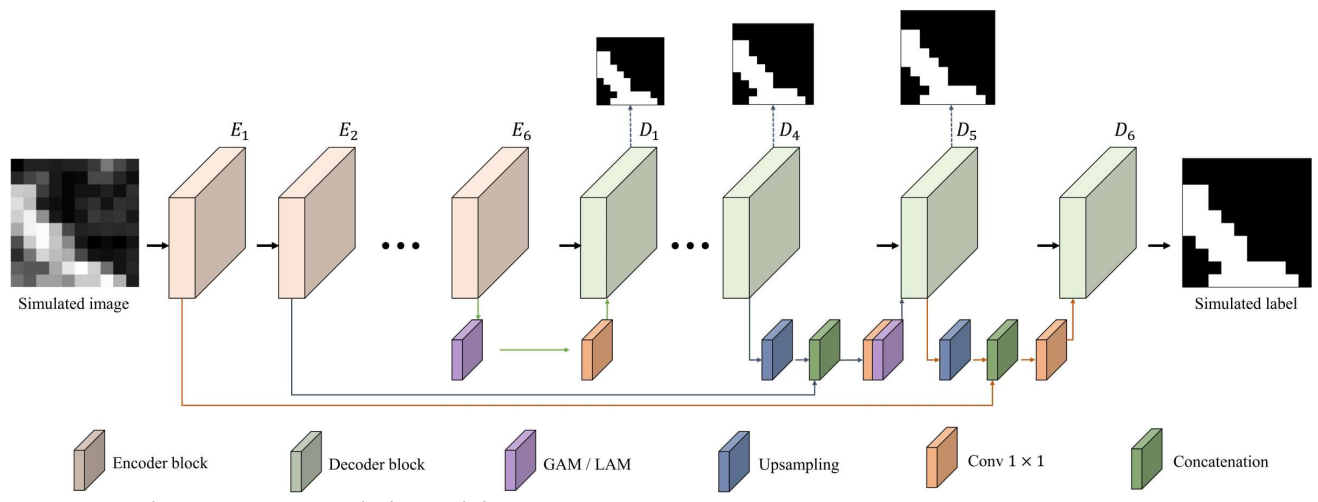

**Fig. 4.** The TDD model architecture.

### *B. Anomaly Sample Simulation*

Anomaly samples are necessary to optimize the model directly. Considering that real anomaly samples are difficult to obtain in real applications, anomaly simulation and generation is needed. The proposed simulation strategy is designed based on the hyperspectral anomaly definition, which stresses that anomaly pixels deviate spectrally from the surroundings [33]. Three useful pieces of information can be drawn from this. Firstly, only the surrounding context is needed to discriminate anomalies. In other words, we can use a patch as the input unit rather than the whole image. This design not only avoids interference from distant pixels, but also increases the sample size. Secondly, a large spectral difference between the anomaly and the background must exist. For each anomaly, the background spectrum present can be used as a spectral deviation reference. Thirdly, the anomaly category is unknown and infinite. Once there is a spectral deviation, it can be seen as an anomaly according to the definition. Therefore, the simulated anomalies must be large in number and not have generation rules that can be easily captured by the model.

Based on the above three points, we designed an anomaly sample simulation strategy consisting of three procedures, an example of which is provided in Fig. 3.

***Procedure 1:*** Anomalous region selection. After cropping the entire training image into patches, this operation is performed for each patch in turn. Because hyperspectral anomalies are generally small targets, this procedure requires control of the anomaly area size when randomly selecting the area so that the background area is larger than the anomaly area. For input patch $\boldsymbol{X}$, the processed patch is denoted as $\boldsymbol{X}_1$. The selected anomaly area is rectangular and masked in $\boldsymbol{X}_1$.

***Procedure 2:*** Anomaly spectrum generation. To create spectral deviation, each pixel of patch $x$ is first randomly shuffled in the spectral dimension to create $\boldsymbol{X}_2$. Assuming that the original spectrum of patch $\boldsymbol{X}$ is the background spectrum (which holds true due to the extremely low proportion of hyperspectral anomalies), the pixels in $\boldsymbol{X}_2$ are all anomalies. Next, cut anomaly pixels at the masked area from Procedure 1 in $\boldsymbol{X}_2$ and paste them into $\boldsymbol{X}_1$ to obtain $\boldsymbol{X}_3$.

***Procedure 3:*** Anomalous region affine transformation. Although $\boldsymbol{X}_3$ is already a trainable anomaly sample, the anomalous regions in $\boldsymbol{X}_3$ are uniformly rectangular, which does not correspond to the real-world situation where anomalies have different shapes. For this reason, random affine transformation is performed on $\boldsymbol{X}_3$ with its corresponding label $\boldsymbol{Y}_3$ to give the anomaly shape information. The applied affine transformation is a combined sequence of rotation, scaling, and translation. For patch $\boldsymbol{X}_3$, the affine transformation process can be formulated as follows:

$$\boldsymbol{X}_4 = \boldsymbol{T}\boldsymbol{X}_3 + b \tag{6}$$

where $\boldsymbol{T}$ is the transformation matrix and $b$ is the translation distance. We let $(c_x, c_y)$ be the center of $\boldsymbol{X}_3$, $\theta$ be the positive anti-clockwise rotation angle, and $s$ be the scale ratio. For simplicity, we define $\alpha = s\cdot\cos(\theta)$, $\beta = s\cdot\sin(\theta)$. $\boldsymbol{T}$ can be expanded as shown in (7):

$$\boldsymbol{T} = \begin{bmatrix} \alpha & \beta & (1-\alpha)\cdot c_x - \beta\cdot c_y \\ -\beta & \alpha & \beta\cdot c_x + (1-\alpha)\cdot c_y \end{bmatrix} \tag{7}$$

In total, the designed anomaly simulation strategy incorporates strong anomaly location randomness, spectral randomness, and shape randomness, under the hyperspectral definition. These properties force the model to learn a "spectrally deviating" relationship rather than a specific background or anomaly target.

### *C. Model Architecture*

To output the anomaly score map directly, thus not relying on any post-processing steps and enabling direct optimization of the HAD task, the proposed architecture is based on the U-Net architecture with encoding and decoding parts (as shown in Fig. 4) [34]. The encoder part contains six cascaded feature extraction blocks for extracting multi-scale, multi-level features. The decoder part contains the decoding blocks corresponding to the encoding blocks in turn. The skip connection between the encoder and decoder parts helps to maintain the important spatial information.

The main architectural innovation is the design of the GAM and the LAM in the decoding part. Because we expect the model to learn the deviating relationship between anomalies and background, rather than being dependent on a specific background, the relationship modeling between pixels in the input image is particularly important. Self-attention mechanisms help with this as they model the correlation between each pixel and the rest of the image pixels [35], avoiding the problem of convolution failing to capture the long-distance dependencies. The GAM calculates the correlation for all the pixels, while the LAM calculates the correlation for pixels in the local range. To reduce the computational burden and suppress local noise, the decoder uses alternating global and local perceptual attention modules. The designed self-attention module can be plug-and-play, without changing the spatial size and feature dimension of the original feature map.

*1) Encoder*

Six cascaded feature extraction blocks form the encoder. Each extraction block consists mainly of several convolutional layers and rectified linear unit (ReLU) activation layers. In view of the small size of the hyperspectral anomaly objects, the convolution kernel is set to $3\times3$ to prevent the loss of anomaly targets during the network forward propagation. Dilated convolution is used in the last block to keep the large size of the feature maps, to avoid losing too many spatial details [36]. As the network layer deepens, the spatial dimension of the extracted feature cube continues to shrink and the feature dimension gradually increases. Changes in the spatial dimensions are implemented using dilation pooling, and the feature dimension changes are implemented using convolution. Finally, the encoder outputs six feature cubes, named $\boldsymbol{E}_1, \boldsymbol{E}_2, ..., \boldsymbol{E}_6$, in sequential order.

*2) Decoder*

The role of the decoder is to decode the features obtained from the encoder into the final anomaly score map. To make full use of the multi-level feature output from the encoder and maintain important spatial details, the decoder is designed as a symmetrical structure having six decoding blocks. The output feature cubes are named $\boldsymbol{D}_1, \boldsymbol{D}_2, ..., \boldsymbol{D}_6$ in sequential order. Each decoding block can be thought of as a function $f$ in (8):

$$\boldsymbol{D}_i = f(\boldsymbol{D}_{i-1}, \boldsymbol{E}_{7-i}) \tag{8}$$

Each decoding block $\boldsymbol{D}_i$ has two inputs: the $\boldsymbol{D}_{i-1}$ from the previous block and the feature cube $\boldsymbol{E}_{7-i}$ corresponding to the output of the encoder. $\boldsymbol{D}_1$ can be seen as a special case where $\boldsymbol{D}_0$ is none. For the subsequent fusion between $\boldsymbol{D}_{i-1}$ and $\boldsymbol{E}_{7-i}$, each $\boldsymbol{D}_{i-1}$ is first interpolated to reach the same spatial dimension as $\boldsymbol{E}_{7-i}$. The fusion step of $f$ can be divided into three sub-steps, as shown in (9)–(10):

$$\boldsymbol{D}_{i1} = \text{Conv}_{1\times1}(\text{Concat}(\boldsymbol{D}_{i-1}, \boldsymbol{E}_{7-i})) \tag{9}$$

$$\boldsymbol{D}_i = \text{GAM/LAM}(\boldsymbol{D}_{i1}) \tag{10}$$

Equation (4) fuses $\boldsymbol{D}_{i-1}$ and $\boldsymbol{E}_{7-i}$ in the channel dimension using $1\times1$ convolution. We let the channel size of $\boldsymbol{D}_{i-1}$ and $\boldsymbol{E}_{7-i}$ be $C$. The convolution in (1) reduces the concatenated feature block channel dimension from $2C$ to $C$. Equation (10) then processes $\boldsymbol{D}_{i1}$ using the designed GAM or LAM, which is elaborated below. The GAM and LAM do not change the feature block size and can achieve a plug-and-play effect. The last decoding block $\boldsymbol{D}_6$ outputs the anomaly score map directly, without the GAM or LAM (i.e., $\boldsymbol{D}_i = \boldsymbol{D}_{i1}$).

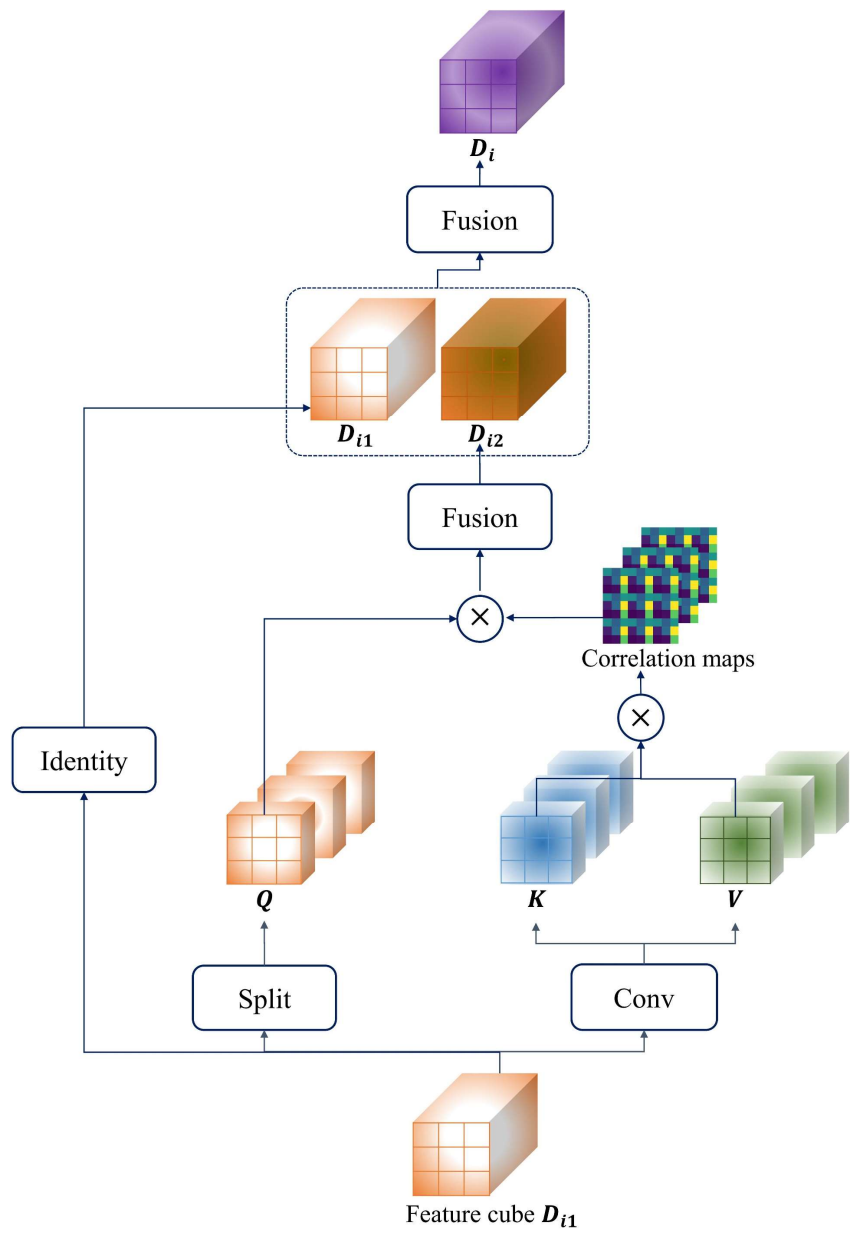

**Fig. 5.** Global self-attention module.

*Global Self-Attention Module.* To make the model focus on the "spectrally deviating" relationship between the anomalies and background, we designed the GAM to explicitly model the pixel correlation in the whole patch. The internal architecture of the GAM is shown in Fig. 5. Inspired by Dosovitskiy *et al.* [37], the query $\boldsymbol{Q}_i$ and key $\boldsymbol{K}_i$ are first generated for $\boldsymbol{D}_{i1}$ using $1\times1$ convolution. Due to the large spectral dimension of the feature cube in the decoder, the generated query and key values are split and thus used in the computation of the multi-head self-attention mechanism. Correspondingly, the same split operation is performed for $\boldsymbol{D}_{i1}$. Assuming that $\boldsymbol{Q}_i$, $\boldsymbol{K}_i$, and $\boldsymbol{D}_{i1}$ are all divided into $n$ segments in the spectral dimension, each segment is named in turn as $\boldsymbol{Q}_i^j$, $\boldsymbol{K}_i^j$, and $\boldsymbol{D}_{i1}^j$ ($1 \le j \le n$). For each combination ($\boldsymbol{Q}_i^j$, $\boldsymbol{K}_i^j$, $\boldsymbol{D}_{i1}^j$), the self-attention mechanism relies on scaled dot-product attention, as given in (11):

$$\text{Attention}(\boldsymbol{Q}_i^j, \boldsymbol{K}_i^j, \boldsymbol{D}_{i1}^j) = \text{softmax}\left(\frac{\boldsymbol{Q}_i^j(\boldsymbol{K}_i^j)^T}{\sqrt{d^k}}\right)\boldsymbol{D}_{i1}^j \tag{11}$$

where $d^k$ is the key dimension and $\text{softmax}\left(\frac{\boldsymbol{Q}_i^j(\boldsymbol{K}_i^j)^T}{\sqrt{d^k}}\right)$ is the obtained correlation map. To make the processed feature cube have the same dimensionality as $\boldsymbol{D}_{i1}$, the $n$ obtained heads ($\text{head}_j = \text{Attention}(\boldsymbol{Q}_i^j, \boldsymbol{K}_i^j, \boldsymbol{D}_{i1}^j), 1 \le j \le n$) are fused as shown in (12), where $W$ is the linear mapping parameter.

$$\boldsymbol{D}_{i2} = \text{Concat}(\text{head}_1, \text{head}_2, ..., \text{head}_n)W \tag{12}$$

$\boldsymbol{D}_{i2}$ contains the fused global pixel correlation information. Considering that $\boldsymbol{D}_{i2}$ may lose the information of the pixel itself due to the introduction of other pixel features, $\boldsymbol{D}_{i2}$ and $\boldsymbol{D}_{i1}$ are further fused using $1\times1$ convolution to obtain the $\boldsymbol{D}_i$ in (10). Equation (13) shows the process where the convolution reduces the concatenated feature dimension to be the same as that of $\boldsymbol{D}_{i1}$.

$$\boldsymbol{D}_i=\text{Conv}_{1\times1}(\text{Concat}(\boldsymbol{D}_{i1},\boldsymbol{D}_{i2})) \tag{13}$$

*Local Self-Attention Module.* Unlike the GAM, the LAM only computes self-attention in local windows, which is more computationally efficient. The LAM can be used as the refinement of the GAM to better maintain local consistency and eliminate the influence of noise on HAD. The internal architecture of the LAM is shown in Fig. 6. We let the size of the local window be $\widehat{H}\times\widehat{W}$ centered at feature $\boldsymbol{x}_c$. For the centered $\boldsymbol{x}_c$, the local perception operation is used to extract the corresponding feature cube $\boldsymbol{F}^c$ with the size $\widehat{H}\times\widehat{W}\times C$. The local correlation operation computes the correlation value between $\boldsymbol{x}_c$ and each remaining feature vector in $\boldsymbol{F}^c$. The obtained correlation map $\boldsymbol{M}^c$ has a size of $\widehat{H}\times\widehat{W}$. For computational efficiency, $\boldsymbol{M}^c$ is calculated using convolution on $\boldsymbol{F}^c$ and the softmax activation function. The processed contextual feature $\boldsymbol{x}_c'$ is then the weighted average of $\boldsymbol{F}^c$ according to $\boldsymbol{M}^c$, as shown in (14):

$$\boldsymbol{x}_c'=\sum_{h=1}^{\widehat{H}}\sum_{w=1}^{\widehat{W}}\boldsymbol{M}_{hw}^c\boldsymbol{F}_{hw}^c \tag{14}$$

where $\boldsymbol{F}_{ij}^c$ is the feature vector in $\boldsymbol{F}^c$ at spatial location $(h,w)$. The above process can be repeated efficiently for all the features in $\boldsymbol{D}_{i1}$ by matrix operations, and then the contextual cube $\boldsymbol{D}_{i2}$ is obtained. Similar to the GAM, the LAM has the same fusion process as shown in (13), after which $\boldsymbol{D}_i$ is finally obtained.

### D. Training Loss

The anomaly sample simulation process generates many pairs of hyperspectral data and anomaly labels, which provide a strong supervised signal for the model training. To make full use of the simulated labels, we add a $1\times1$ convolutional layer and sigmoid activation layer on top of each $\boldsymbol{D}_i(1\le i\le5)$ to generate anomaly maps $\boldsymbol{S}_i$, as shown in (15):

$$\boldsymbol{S}_i=\text{Sigmoid}(\text{Conv}_{1\times1}(\boldsymbol{D}_i)) \tag{15}$$

where $i=6$ and $\boldsymbol{S}_6=\boldsymbol{D}_6$. The six generated $\boldsymbol{S}_i$ have different spatial sizes, keeping the same size as their corresponding $\boldsymbol{D}_i$. Finally, the ground truth $\boldsymbol{G}$ is resized to the same size for each $\boldsymbol{D}_i$, denoted as $\boldsymbol{G}_i$ with spatial size $H_i\times W_i$, to optimize the model based on the weighted average cross-entropy (CE) loss $L$:

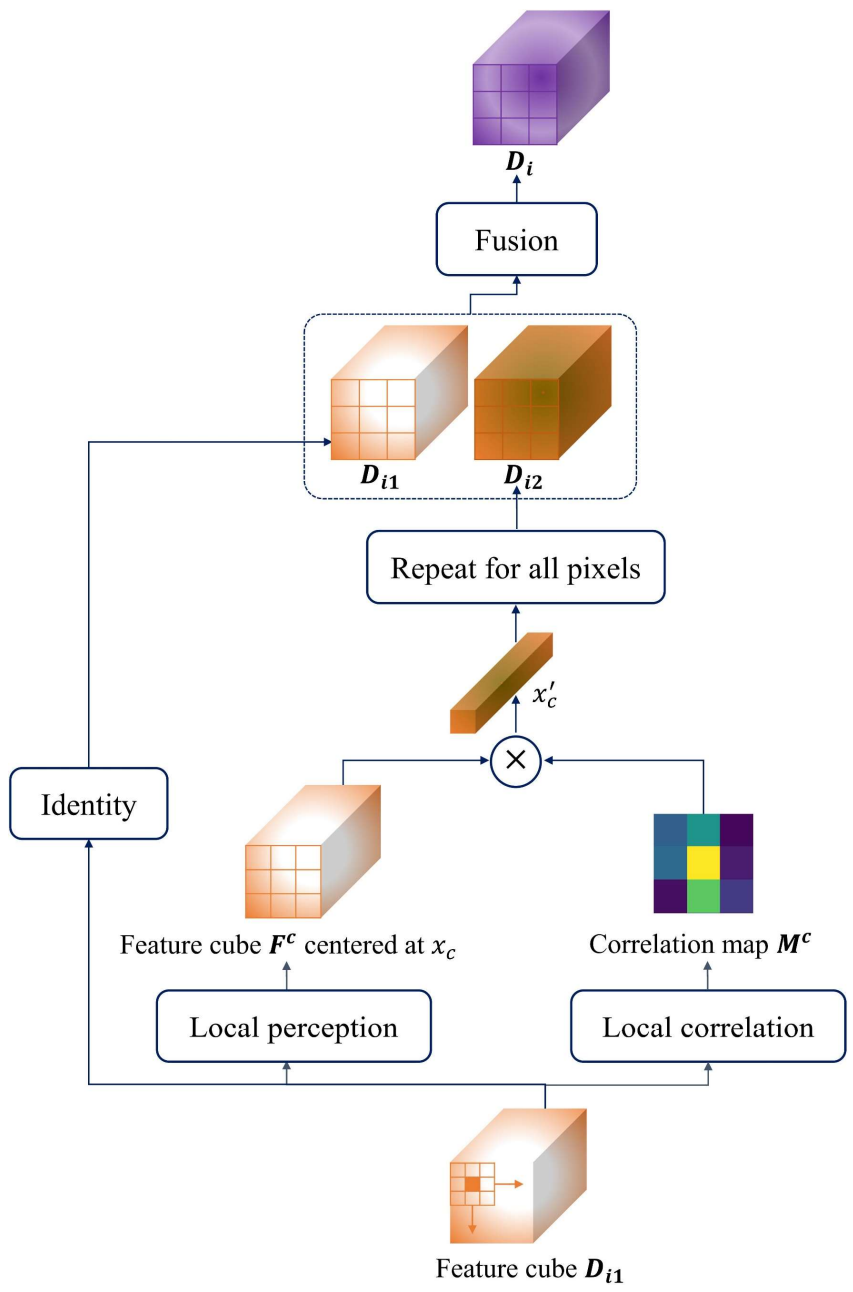

**Fig. 6.** Local self-attention module.

$$L=-\sum_{i=1}^{6}\omega_i\sum_{h=1}^{H_i}\sum_{w=1}^{W_i}\begin{matrix}[\boldsymbol{G}_i(h,w)\log\boldsymbol{S}_i(h,w)\\+(1-\boldsymbol{G}_i(h,w))\log(1-\boldsymbol{S}_i(h,w))]\end{matrix} \tag{16}$$

where $\omega_i$ is the weight for the CE loss of $\boldsymbol{G}_i$ and $\boldsymbol{S}_i$.

### E. Model Transferability

Unlike the previous deep models that can only train and reason on a single image, the proposed TDD model has the ability to migrate between different images. In other words, we only need to train once on a single image to infer on many other unseen images. This is because the TDD model is required to learn the "spectrally deviating" properties of anomalies, rather than the specific background, as described in detail in sections A, B, and C.

Despite this, there is still a remaining problem to be solved when the TDD model infers between different data. The input layer of the built network architecture requires a fixed number of bands, but different numbers of bands for different data. To solve this problem, we need to process the data in terms of the channel dimensions, before inference. We let the number of channels of the training data be $B_1$ and the number of channels of the test data be $B_2$. If $B_2<B_1$, bilinear interpolation of the test data is performed in the spectral dimension to achieve a spectral dimension of $B_1$. If $B_2>B_1$, the test data are cut into many segments along the spectral dimension, where the spectral dimension of each segment is $B_1$. When the spectral dimension of the last segment is less than $B_1$, the $B_1$ bands of the last segment are taken from the end of the spectrum. Finally, the mean of all the segment detection results is used as the final output.

It is worth noting that the whole process does not abandon any of the original bands. The above trick that is used to deal

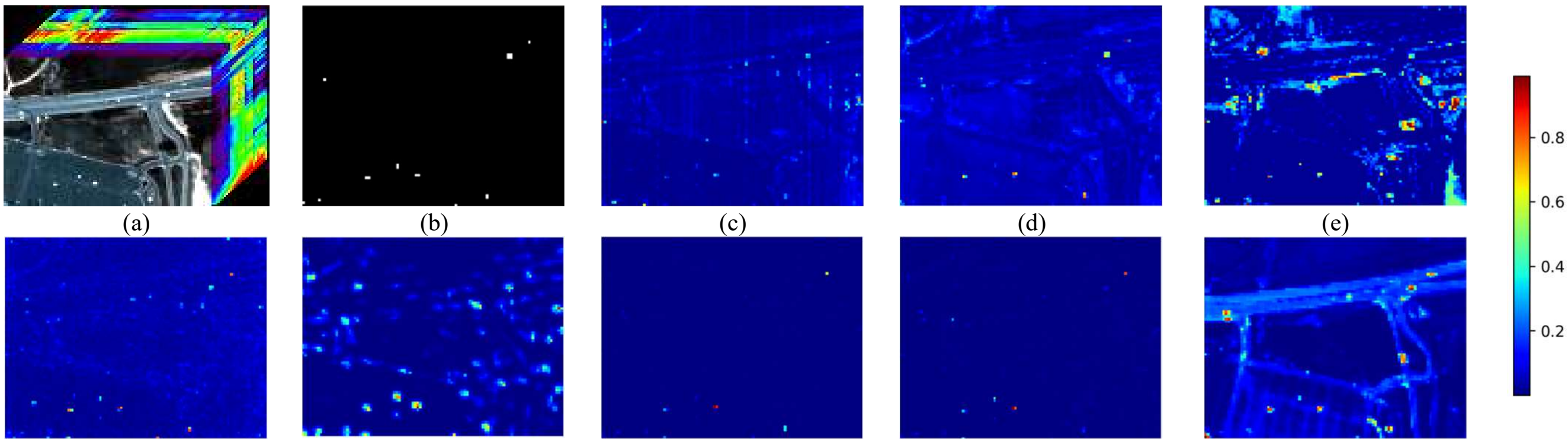

**Fig. 7.** The anomaly detection results obtained on the HYDICE dataset. (a) Data cube. (b) The ground-truth map. (c) GRX. (d) LRASR. (e) ADLR. (f) CRD. (g) SC_AAE. (h)Auto-AD. (i) DeepLR. (j) TDD.

with the varying image channels is simple but effective in practice.

## IV. Experiments and Analysis

### A. Experimental Settings

#### 1) Datasets

In the experiments, the proposed TDD model was validated on four hyperspectral datasets: 1) a HYDICE dataset; 2) the Airborne Visible-Infrared Imaging Spectrometer (AVIRIS)-1 dataset; 3) the Cri dataset; and 4) the WHU-Hi-River dataset.

The HYDICE dataset was collected by the HYDICE sensor [38]. The dataset contains 162 spectral bands covering 400–2500 nm. The spectral resolution is 10 nm and the spatial resolution is 1 m/pixel. The image size is 80×100 pixels, and 10 vehicles are marked as anomalies (occupying 17 pixels).

The AVIRIS-1 dataset [39] was captured by the AVIRIS sensor in San Diego, CA, USA. The imagery contains 186 spectral bands covering 400–2500 nm. The AVIRIS-1 dataset has 100 × 100 pixels, and three planes are marked as anomalies (occupying 143 pixels). The spatial resolution is 3.5 m/pixel.

The Cri dataset [30] was acquired by the Nuance Cri hyperspectral sensor, with the spectral resolution of 10 nm. The image has 46 spectral bands in the wavelength range of 650–1100 nm, and 400×400 pixels. The rocks are considered as the anomalies, occupying 1254 pixels.

The WHU-Hi-River dataset [39] was captured by a Headwall Nano-Hyperspec sensor mounted on a unmanned aerial vehicle (UAV) platform, and has a spatial size of 105×168 pixels. The image has 135 spectral bands covering the spectral range of 400–1000 nm. The spatial resolution is 6 cm/pixel. Two plastic plates and two gray panels are considered as the anomalies in the anomaly detection task, with a total number of 36 pixels.

We trained the TDD model using the simulated anomaly samples from the HYDICE dataset and then inferred the model directly on all four datasets. We chose the HYDICE dataset as the training set because its background contains more homogeneous categories, and tiny anomalies bring less contamination to the simulated labels.

#### 2) Comparison Models and Evaluation Metrics

The TDD model was compared with the following seven models: the global RX detector (GRX) [8], the low-rank and sparse matrix decomposition-based Mahalanobis distance (LRASR) method [40], the abundance and dictionary-based low-rank decomposition (ADLR) [41] method, the collaborative-representation-based (CRD) method [42], the spectral constraint autoencoders (SC_AAE) method [18], the autonomous hyperspectral anomaly detection network based on a fully convolutional AE (Auto-AD) [15], and the deep low-rank prior-based method (DeepLR) [1]. The comparison methods cover the three categories of RXD-based, representation-based, and deep learning based methods, and all of them are classic algorithms. Among the different methods, Auto-AD and DeepLR are the very recently proposed state-of-the-art models.

In this paper, multi-parameter 3D receiver operating characteristic (3D ROC) curves are used for the quantitative evaluation [1]. Compared to 2D ROC curves, 3D ROC curves add the threshold dimension. There are three basic area under the curve (AUC) values that can be derived from the 3D ROC curves: the AUC score of the ROC curve $(P_D, t)$ ($AUC_{(D,t)}$), the AUC score of the ROC curve $(P_F, t)$ ($AUC_{(F,t)}$), and the AUC score of the ROC curve ( $P_D$ , $P_F$ ) ($AUC_{(D,F)}$), where $P_D$ is the probability of detection, $P_F$ is the false alarm rate, and $t$ is the threshold. Based on the three AUC values, four measures can be further derived, i.e., the target detectability of a detector ($AUC_{TD}$), the background suppressibility of a detector ($AUC_{BS}$), the overall detection probability ($AUC_{ODP}$), and the signal-to-noise probability ratio ($AUC_{SNPR}$). A larger $AUC_{(D,t)}$, $AUC_{(D,F)}$, $AUC_{TD}$, or $AUC_{ODP}$ means a better detection performance. A smaller $AUC_{(F,t)}$ or larger $AUC_{BS}$ means a better background suppression performance. $AUC_{SNPR}$ demonstrates the signal-to-noise probability ratio of the obtained detection map.

#### 3) Implementation Details

We set the cluster number and the selected pixels of LRASR to 15 and 20, according to the original paper. The regularization parameter of CRD was set to $10^{-6}$. The CRD model adopted a dual-window strategy with the inner window size $w_{in}$ and the outer window size $w_{out}$. $(w_{in}, w_{out})$ were set to (7,15) for the HYDICE dataset and the WHU-Hi-River dataset, and (11,17) for the AVIRIS-1 dataset and the Cri dataset. The U-Net model was adopted to achieve the feature cube for the SC_AAE model. The threshold of DeepLR was set to 0.00001 for the HYDICE dataset and 0.0001 for the other datasets. The GAM and LAM were used alternately in the decoder part of the TDD model. For the first five decoding blocks, the order of use was LAM-GAM-LAM-GAM-LAM. For the HYDICE, AVIRIS-1, Cri, and WHU-Hi-River datasets, the patch size for the training and inference was set to 10, 50, 100, and 20, respectively. $\omega_1, \omega_2$, and $\omega_3$ were set to 0.5. $\omega_4, \omega_5$, and $\omega_6$ were set to 1.0. The CPU was an Intel(R) Xeon(R) CPU E5-2690 v4 @ 2.60 GHz with 62 GB memory, and the GPU was a Tesla P100-PCIE with 16 GB of memory.

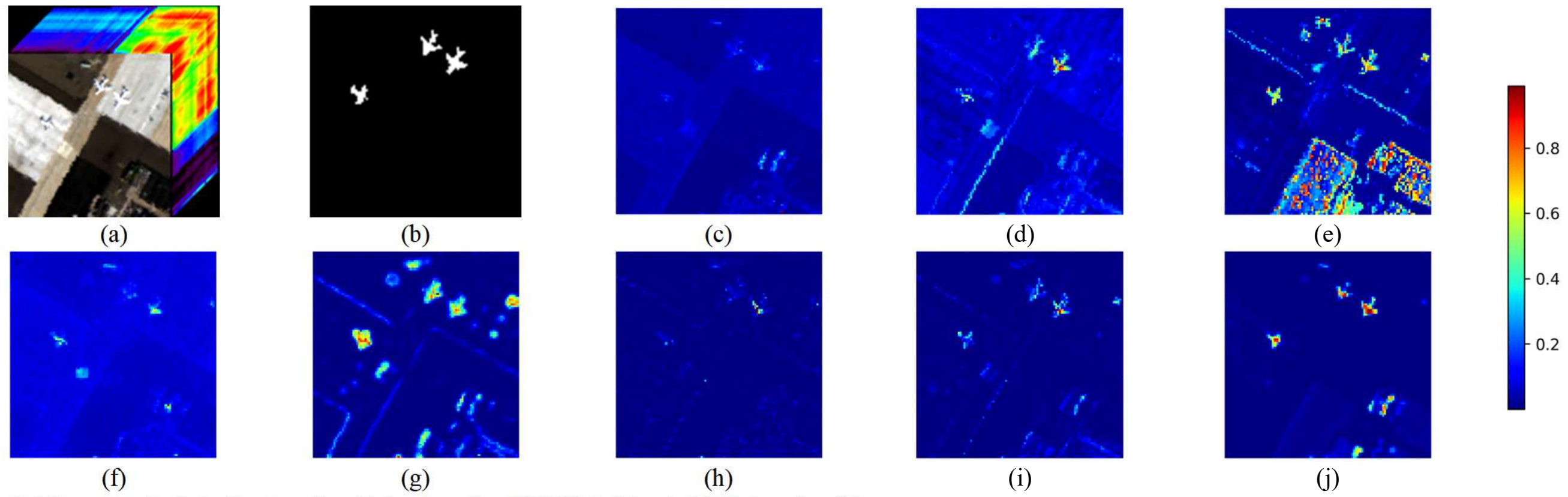

**Fig. 8.** The anomaly detection results obtained on the AVIRIS-1 dataset. (a) Data cube. (b) The ground-truth map. (c) GRX. (d) LRASR. (e) ADLR. (f) CRD. (g) SC_AAE. (h)Auto-AD. (i) DeepLR. (j) TDD.

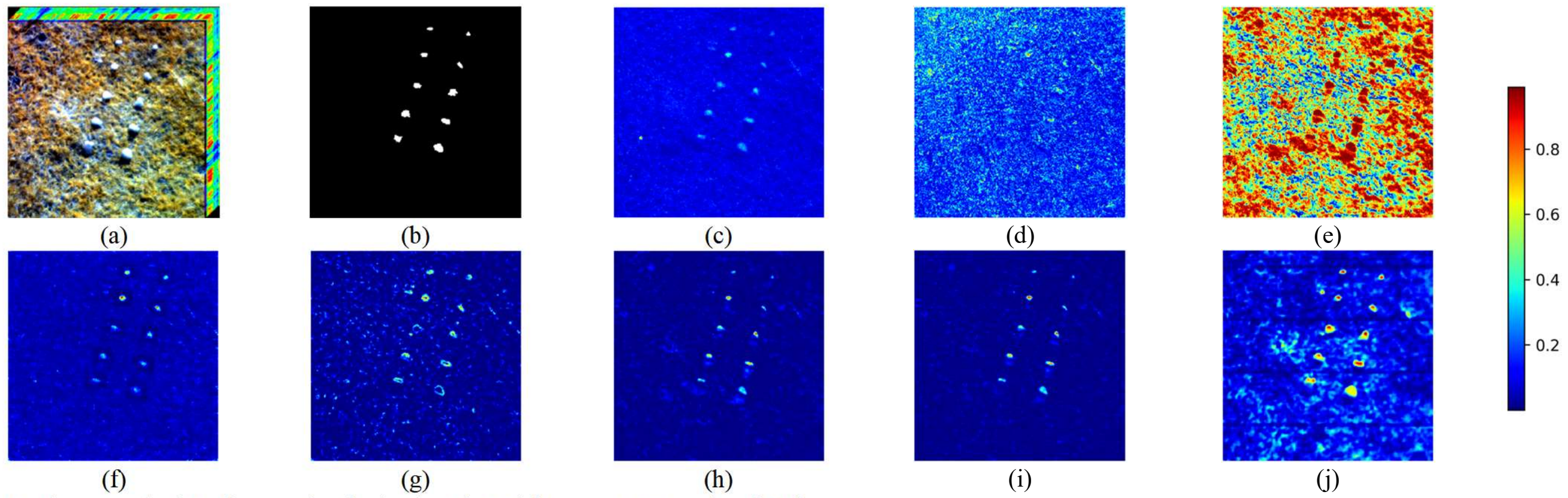

**Fig. 9.** The anomaly detection results obtained on the Cri dataset. (a) Data cube. (b) The ground-truth map. (c) GRX. (d) LRASR. (e) ADLR. (f) CRD. (g) SC_AAE. (h)Auto-AD. (i) DeepLR. (j) TDD.

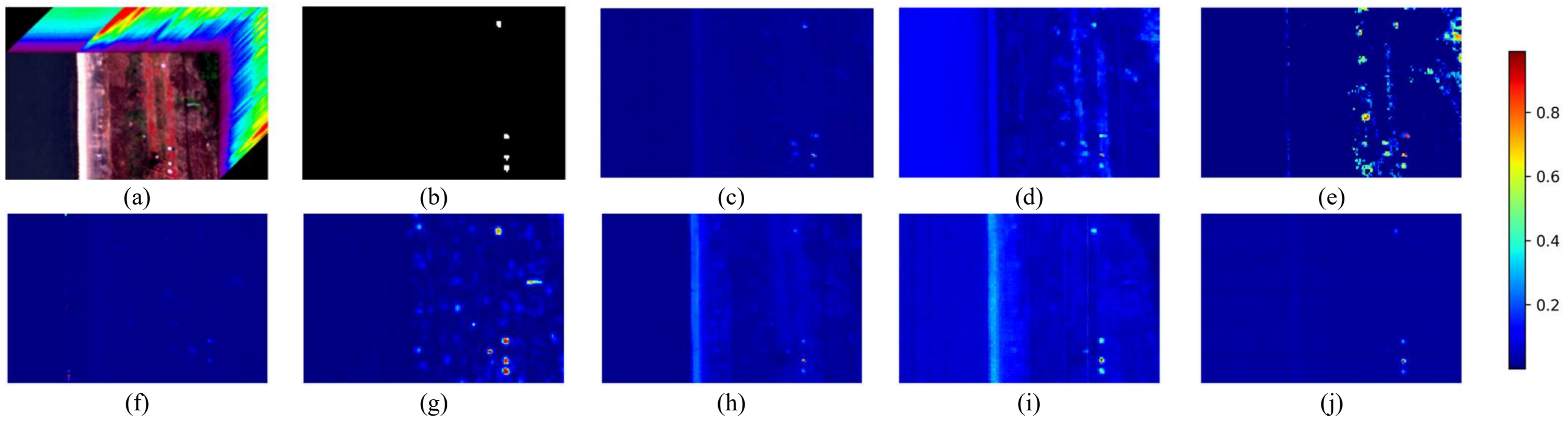

**Fig. 10.** The anomaly detection results obtained on the WHU-Hi-River dataset. (a) Data cube. (b) The ground-truth map. (c) GRX. (d) LRASR. (e) ADLR. (f) CRD. (g) SC_AAE. (h)Auto-AD. (i) DeepLR. (j) TDD.

### B. *Qualitative Evaluation*

In this section, we display the obtained anomaly maps for the eight comparison algorithms on the four datasets. It is worth noting that the TDD model was trained on the HYDICE dataset in an unsupervised manner and directly inferred on the remaining three datasets, without any further fine-tuning.

#### *1) HYDICE Dataset*

Fig. 7 shows the obtained anomaly maps for the HYDICE dataset. In the anomaly maps obtained by GRX, LRASR, and CRD, the brightness difference between the anomalies and the background is small and visually indistinguishable. The anomaly maps of ADLR and SC_AAE have many false alarms, and the brightness of the false alarms and anomalies are at the same level. Although Auto-AD and DeepLR have low false alarm rates, some anomalies cannot be visually identified. In contrast, in the anomaly map obtained by the TDD model, the anomaly points have a greater difference with the background, and the false alarm rate is lower than for ADLR and SC_AAE.

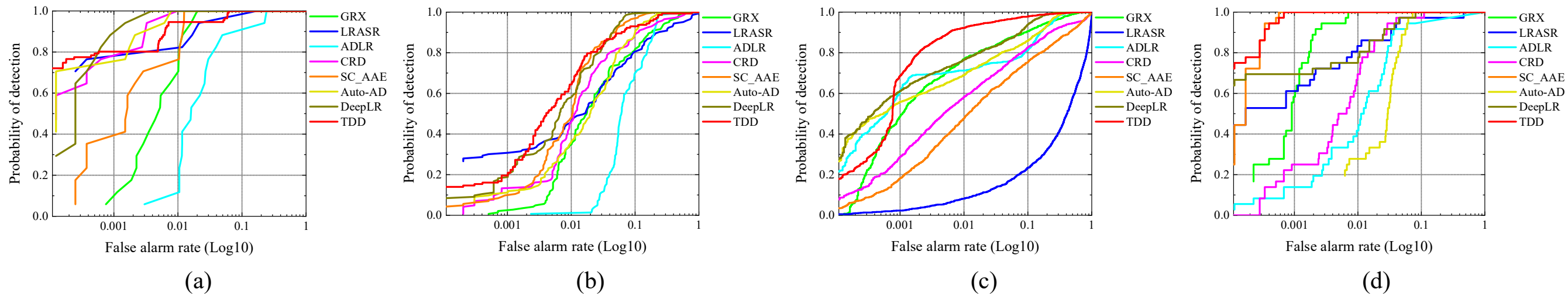

**Fig. 11.** 2D ROC curves of $(P_D, P_F)$. (a) HYDICE dataset. (b) AVIRIS-1 dataset. (c) Cri dataset. (d) WHU-Hi-River dataset.

*2) AVIRIS-1 Dataset*

Fig. 8 shows the obtained anomaly maps for the AVIRIS-1 dataset. Among the eight comparison algorithms, some of the algorithms cannot fully identify the aircraft, including LRASR, CRD, Auto-AD, and DeepLR. In the anomaly map of GRX, the response value of the abnormal area is low and close to that of the background. Although ADLR and SC_AAE can identify more complete aircraft, there are high false alarm rates. In contrast, the TDD model shows more complete anomaly identification and a lower false alarm rate.

*3) Cri Dataset*

Fig. 9 shows the obtained anomaly maps for the Cri dataset. In the anomaly map of GRX, the contour of the anomaly area is close to the ground truth, but the response value is lower. LRASR and ADLR show poor detection performances, with anomalies and background almost indistinguishable. CRD, SC_AAE, Auto-AD, and DeepLR can all locate the anomalous targets, but the detected areas are incomplete. In contrast, the TDD model can detect the anomalous areas relatively completely. Although some background areas have higher brightness values, there is still a big difference with the brightness of the anomalous areas.

*4) WHU-Hi-River Dataset*

Fig. 10 shows the obtained anomaly maps for the WHU-Hi-River dataset. The anomaly maps of LRASR, ADLR, and SC_AAE have high false alarm rates, and the anomaly maps of LRASR and ADLR cannot distinguish between anomalies and false alarms. The overall response value of the detection map of CRD is low, and the abnormal area cannot be visually interpreted. GRX, Auto-AD, DeepLR, and TDD can all visually discriminate the abnormal areas, but the TDD model shows a better background suppression effect.

## C. *Quantitative Evaluation*

This section provides the quantitative evaluation results for the four datasets, including the ROC curves $(P_D, P_F)$ in Fig. 11, and the separability maps in Fig. 12. The AUC scores are listed in Tables I–IV, where the bold figures represent the best results in the corresponding metrics.

*1) HYDICE Dataset*

The ROC curves $(P_D, P_F)$ are shown in Fig. 11(a), and the AUC

TABLE I
AUC Scores for the HYDICE Dataset

| Method | $AUC_{(D,F)}$ | $AUC_{(D,\tau)}$ | $AUC_{(F,\tau)}$ | $AUC_{TD}$ | $AUC_{BS}$ | $AUC_{ODP}$ | $AUC_{SNPR}$ |
|---|---|---|---|---|---|---|---|
| GRX | 0.9938 | 0.2487 | 0.0571 | 1.2425 | 0.9367 | 1.1916 | 4.3555 |
| LRASR | 0.9920 | 0.5189 | 0.0490 | 1.5109 | 0.9430 | 1.4699 | 10.5898 |
| ADLR | 0.9624 | 0.4640 | 0.0713 | 1.4264 | 0.8911 | 1.3927 | 6.5077 |
| CRD | 0.9991 | 0.5145 | 0.0576 | 1.5136 | 0.9424 | 1.4578 | 9.0741 |
| SC_AAE | 0.9962 | 0.5458 | 0.1487 | 1.5420 | 0.8475 | 1.3971 | 3.6705 |
| Auto-AD | 0.9991 | 0.2756 | **0.0070** | 1.2747 | **0.9921** | 1.2686 | **39.3714** |
| DeepLR | **0.9996** | 0.3054 | 0.0156 | 1.3050 | 0.9840 | 1.2898 | 19.5769 |
| TDD | 0.9960 | **0.8757** | 0.0660 | **1.8717** | 0.9300 | **1.8057** | 13.2682 |

TABLE II
AUC Scores for the AVIRIS-1 Dataset

| Method | $AUC_{(D,F)}$ | $AUC_{(D,\tau)}$ | $AUC_{(F,\tau)}$ | $AUC_{TD}$ | $AUC_{BS}$ | $AUC_{ODP}$ | $AUC_{SNPR}$ |
|---|---|---|---|---|---|---|---|
| GRX | 0.9370 | 0.0968 | 0.0309 | 1.0338 | 0.9061 | 1.0659 | 3.1327 |
| LRASR | 0.9146 | 0.2956 | 0.0665 | 1.2102 | 0.8481 | 1.2291 | 4.4451 |
| ADLR | 0.9081 | 0.3997 | 0.0852 | 1.3078 | 0.8229 | 1.3145 | 4.6913 |
| CRD | 0.9530 | 0.1857 | 0.0686 | 1.1387 | 0.8844 | 1.1171 | 2.7070 |
| SC_AAE | 0.9820 | **0.4607** | 0.0307 | **1.4427** | 0.9513 | **1.4300** | 15.0065 |
| Auto-AD | 0.9628 | 0.0884 | **0.0053** | 1.0512 | 0.9575 | 1.0831 | 16.6792 |
| DeepLR | **0.9845** | 0.2013 | 0.0098 | 1.1858 | **0.9747** | 1.1915 | 20.5408 |
| TDD | 0.9728 | 0.3741 | 0.0093 | 1.3469 | 0.9635 | 1.3376 | **40.2259** |

TABLE III
AUC Scores for the Cri Dataset

| Method | $AUC_{(D,F)}$ | $AUC_{(D,\tau)}$ | $AUC_{(F,\tau)}$ | $AUC_{TD}$ | $AUC_{BS}$ | $AUC_{ODP}$ | $AUC_{SNPR}$ |
|---|---|---|---|---|---|---|---|
| GRX | 0.9678 | 0.2254 | 0.0896 | 1.1932 | 0.8782 | 1.1036 | 2.5156 |
| LRASR | 0.8652 | 0.2327 | 0.1908 | 1.0979 | 0.6744 | 0.9071 | 1.2196 |
| ADLR | 0.9579 | **0.9674** | 0.6420 | **1.9253** | 0.3159 | 1.2833 | 1.5068 |
| CRD | 0.9186 | 0.2164 | 0.0448 | 1.1350 | 0.8738 | 1.0902 | 4.8303 |
| SC_AAE | 0.8849 | 0.2506 | 0.0241 | 1.1355 | 0.8608 | 1.1114 | 10.3983 |
| Auto-AD | 0.9643 | 0.2496 | 0.0150 | 1.2139 | 0.9493 | 1.1989 | 16.6400 |
| DeepLR | 0.9815 | 0.2650 | **0.0128** | 1.2465 | **0.9687** | 1.2337 | **20.7031** |
| TDD | **0.9915** | 0.6383 | 0.1122 | 1.6298 | 0.8793 | **1.5176** | 5.6889 |

TABLE IV
AUC Scores for the WHU-Hi-River Dataset

| Method | $AUC_{(D,F)}$ | $AUC_{(D,\tau)}$ | $AUC_{(F,\tau)}$ | $AUC_{TD}$ | $AUC_{BS}$ | $AUC_{ODP}$ | $AUC_{SNPR}$ |
|---|---|---|---|---|---|---|---|
| GRX | 0.9988 | 0.1999 | 0.0211 | 1.1987 | 0.9777 | 1.1776 | 9.4739 |
| LRASR | 0.9815 | 0.4090 | 0.0682 | 1.3905 | 0.9133 | 1.3223 | 14.3915 |
| ADLR | 0.9560 | 0.3955 | 0.0140 | 1.3515 | 0.9420 | 1.3375 | 68.2857 |
| CRD | 0.9885 | 0.0449 | **0.0083** | 1.0334 | **0.9802** | 1.0251 | **119.0963** |
| SC_AAE | 0.9998 | **0.8144** | 0.0522 | **1.8142** | 0.9476 | **1.7620** | 19.1532 |
| Auto-AD | 0.9725 | 0.2467 | 0.0373 | 1.2192 | 0.9352 | 1.1819 | 26.0724 |
| DeepLR | 0.9925 | 0.4692 | 0.0776 | 1.4617 | 0.9149 | 1.3841 | 12.7899 |
| TDD | **0.9999** | 0.3985 | 0.0279 | 1.3984 | 0.9720 | 1.3705 | 35.8387 |

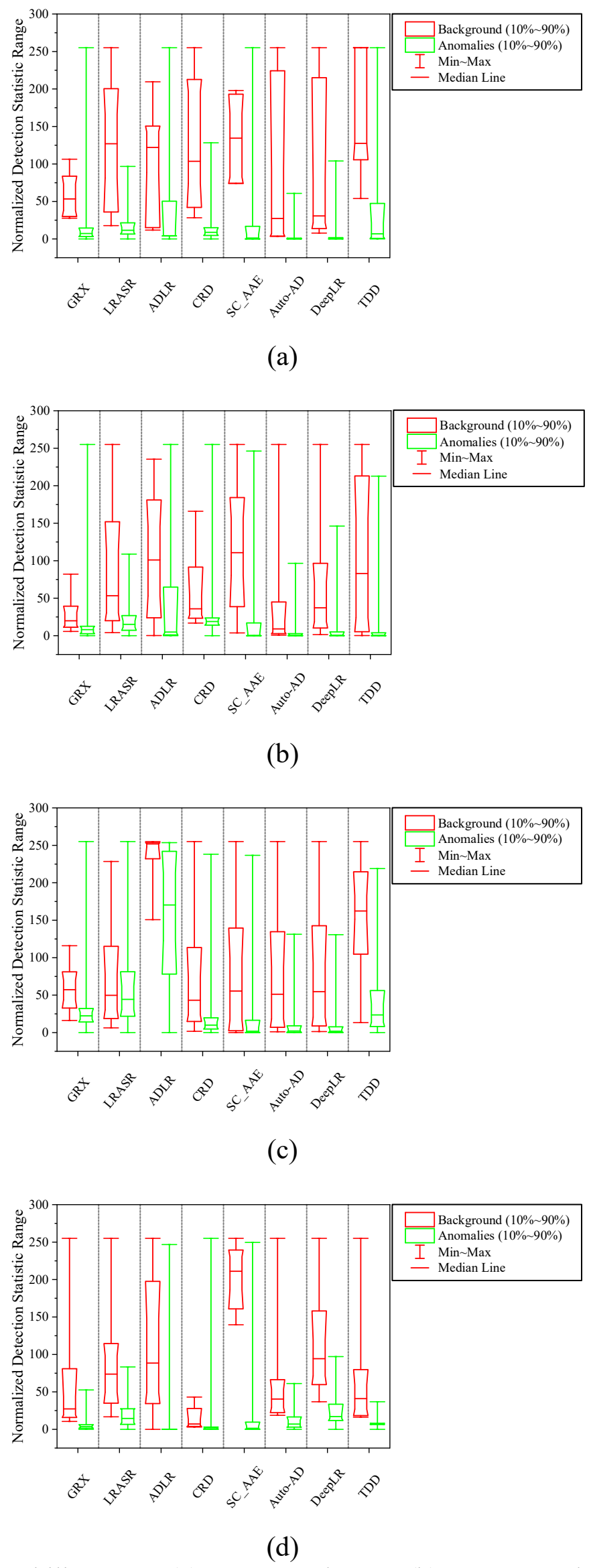

**Fig. 12.** Separability maps. (a) HYDICE dataset. (b) AVIRIS-1 dataset. (c) Cri dataset. (d) WHU-Hi-River dataset.

scores are listed in Table I. The TDD model achieves the best results in $\text{AUC}_{(D,\tau)}$, $\text{AUC}_{TD}$, and $\text{AUC}_{ODP}$, and surpasses the second best performing model by 0.6270, 1.6292, and 0.6141, respectively. Except for ADLR, the $\text{AUC}_{(D,F)}$ scores of most of the models on the HYDICE dataset exceed 0.99, and the models are at the same level.

The separability maps are shown in Fig. 12(a). Compared with the other models, TDD and SC_AAE show a larger difference between the background and anomalies, which is consistent with the visualized anomaly maps in Fig. 6. In addition, the TDD model is the only model with a large number of background response values close to 250 in the separability map, which reflects the ability of the TDD algorithm to learn the spectral deviation, rather than a background distribution, to a certain extent.

### *2) AVIRIS-1 Dataset*

The ROC curves $(P_D, P_F)$ are shown in Fig. 11(b), and the AUC scores are listed in Table II. The TDD model is in the upper-middle level in both the AUC score and ROC curve. Although the TDD model was directly inferred on the AVIRIS-1 dataset, both the $\text{AUC}_{TD}$ and $\text{AUC}_{ODP}$ metrics are second only to SC_AAE, by 0.0958 and 0.0924, respectively, with only a small gap. For the $\text{AUC}_{SNPR}$ indicator, the TDD model achieves the best result of 40.2259, far exceeding the second-best performing model result of 19.6851.

The separability maps are shown in Fig. 12(b). In the results of GRX, LRASR, ADLR, CRD, and Auto-AD, there is a crossover between the background and abnormal body distribution. Consistent with the AUC score results, the SC_AAE model shows better separation, and TDD and DeepLR are at suboptimal levels.

### *3) Cri Dataset*

The ROC curves $(P_D, P_F)$ are shown in Fig. 11(c), and the AUC scores are listed in Table III. Although the TDD model was directly reasoned on the Cri dataset, it still achieves the highest $\text{AUC}_{(D,F)}$ score, far exceeding the second best performing model of DeepLR by 0.01 points, and exceeds 0.99 on this dataset for the first time. The TDD model surpasses the suboptimal ADLR model by 0.2343 points in $\text{AUC}_{ODP}$, reflecting the strong transferability of the TDD model.

The separability maps are shown in Fig. 12(c). The TDD model is the only model that achieves a clear separation effect between background and anomalies, and performs far better than the comparison models.

### *4) WHU-Hi-River Dataset*

The ROC curves $(P_D, P_F)$ are shown in Fig. 11(d), and the AUC scores are listed in Table IV. The TDD model achieves the highest $\text{AUC}_{(D,F)}$ score of 0.999, which is very close to 1.0. However, the TDD model is in the middle of the pack in terms of $\text{AUC}_{(D,\tau)}$, $\text{AUC}_{TD}$, and $\text{AUC}_{ODP}$, being surpassed by the best-performing SC_AAE model by some margin. The TDD model is in the upper-middle level in the $\text{AUC}_{(F,\tau)}$, $\text{AUC}_{BS}$, and $\text{AUC}_{SNPR}$ metrics.

TABLE V
ABLATION EXPERIMENTS FOR THE ATTENTION MODULE IN THE DECODER

| Settings | HYDICE | AVIRIS-1 | Cri | WHU-Hi-River |
|---|---|---|---|---|
| GGGGG | 0.9925 | **0.9740** | 0.9494 | 0.9664 |
| LLLLL | 0.9936 | 0.8916 | **0.9981** | 0.9831 |
| LGLGL | **0.9960** | 0.9728 | 0.9915 | **0.9999** |
| No (L/G) | 0.9911 | 0.9024 | 0.8706 | 0.9846 |

TABLE VI
THE $\text{AUC}_{(D,F)}$ RESULTS OF TRAINING THE TDD MODEL ON DIFFERENT DATASETS

| Training image | HYDICE | AVIRIS-1 | Cri | WHU-Hi-River |
|---|---|---|---|---|
| HYDICE | **0.9960** | 0.9728 | 0.9915 | **0.9999** |
| AVIRIS-1 | 0.9744 | **0.9914** | 0.9855 | 0.9770 |
| Cri | 0.9952 | 0.7353 | **0.9997** | 0.9777 |
| WHU-Hi-River | 0.8588 | 0.5740 | 0.9678 | 0.9996 |

The separability maps are shown in Fig. 12(d). The TDD model achieves a relatively small splitting distance, despite the separation of the background and anomalous subject parts. This explains why the TDD model scores highly in the $AUC_{(D, F)}$ metric, but mid-range in the $AUC_{(D, \tau)}$, $AUC_{TD}$, and $AUC_{ODP}$ metrics. However, the TDD model can achieve an $AUC_{(D, F)}$ score of 0.999 just by direct inference, which demonstrates its superior performance.

### *D. Model Analysis*

In this section, we describe the ablation experiments conducted with the LAM and GAM, and their alternating use. Furthermore, we explore the effect of different training data on the transferability of the TDD model.

Table V lists the results of the ablation experiments, where the bold figures represent the best results in the corresponding datasets. The model setting represents the case of how the GAM (G for short) or LAM (L for short) is used in the decoder. Compared with the case of no attention module, both the GAM and LAM can significantly increase the transferability of the TDD model. Compared with using only one kind of attention mechanism (i.e., the case of only the GAM or LAM), the alternating use of the LAM and GAM can further improve the detection ability.

Table VI lists the $AUC_{(D, F)}$ scores of training the TDD model on different datasets. When choosing to train the TDD model on the HYDICE dataset and the AVIRIS-1 dataset, the transferred $AUC_{(D, F)}$ results are both above 0.9770, showing the high transferability. However, the transfer suffers from instability issues after training on the Cri dataset and WHU-Hi-River dataset, especially when testing on the AVIRIS-1 dataset. We speculate that this may be due to the noisier simulated labels that were generated in the Cri dataset and the WHU-Hi-River dataset. Furthermore, the results listed in Table VI show that, if the transferability is not considered, a higher $AUC_{(D, F)}$ score can be achieved (0.9914 on the AVIRIS-1 dataset and 0.9997 on the Cri dataset) by training and inferring the TDD model on a single dataset.

## V. Conclusion

In this paper, a one-step HAD model called the TDD model has been proposed, which is optimized directly for the HAD task, rather than adopting a proxy task as in traditional models. The spectral deviation relationship is focused in TDD model, which guarantees the transferability and overcomes the limitation of traditional model training and testing on a single image. To train the TDD model in an unsupervised manner, an anomaly sample simulation strategy is proposed to generate numerous anomaly samples based on the hyperspectral anomaly definition. To help the model focus on the context relationship features, we designed the GAM and LAM modules to learn the deviating context explicitly. The TDD model was validated on four public datasets, which confirmed the powerful detection ability and excellent transferability of the model. To the best of our knowledge, the TDD model is the first unsupervised HAD model based on the one-step detection paradigm.